\documentclass[letterpaper, 10 pt, conference]{ieeeconf}  

\usepackage{graphics} 
\usepackage{times} 
\usepackage{amsmath} 
\usepackage{amssymb}  

\usepackage{booktabs}
\usepackage{booktabs}       
\usepackage{pifont}
\usepackage[format=plain,labelsep=period,font=footnotesize,compatibility=false]{caption}

\usepackage[table,dvipsnames]{xcolor}
\usepackage{makecell}

\usepackage{tabu}
\usepackage{multirow}
\usepackage{soul}  
\usepackage[export]{adjustbox}

\usepackage{microtype}
\usepackage{inconsolata}
\usepackage{accents}
\usepackage[symbol]{footmisc}

\newcommand{\vect}[1]{\accentset{\rightharpoonup}{#1}}

\newcommand{\variable}[1]{\texttt{#1}}

\newcommand{\TP}{\textit{TP}}
\newcommand{\FP}{\textit{FP}}

\newcommand{\FN}{\textit{FN}}

\usepackage[capitalize]{cleveref}
\crefname{section}{Sec.}{Secs.}
\Crefname{section}{Section}{Sections}
\Crefname{table}{Table}{Tables}
\crefname{table}{Tab.}{Tabs.}

\usepackage[noadjust]{cite}

\title{\LARGE \bf
LET-3D-AP: Longitudinal Error Tolerant 3D Average Precision for Camera-Only 3D Detection
}

\author{Wei-Chih Hung \qquad Vincent Casser \qquad Henrik Kretzschmar$^{*}$  \qquad Jyh-Jing Hwang  \qquad Dragomir Anguelov  \\
Waymo LLC \\
\vspace{-6mm}
}

\begin{document}

\twocolumn[{%
\renewcommand\twocolumn[1][]{#1}%
\vspace{-4mm}
\maketitle
\vspace{-8mm}
\begin{center}
  \begin{tabular}{c}
  \includegraphics[height=3.8cm]{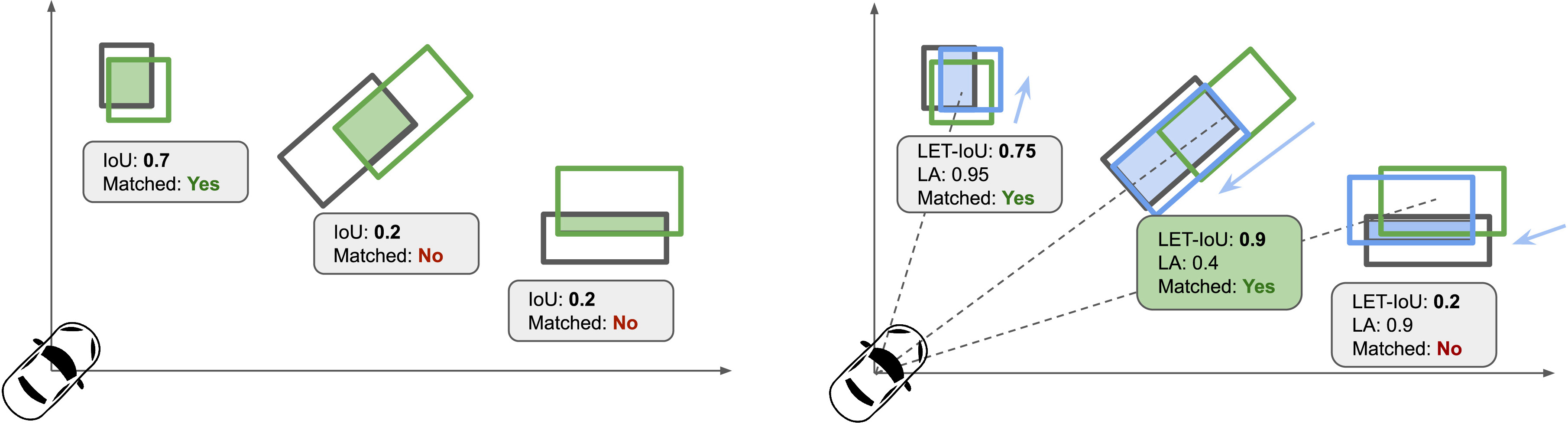} 
  \end{tabular}
\captionof{figure}{\textbf{Evaluating camera-only 3D detections when using the 3D~Average Precision (3D\,AP) metric (left) and when using the proposed longitudinal error tolerant, LET-3D-AP(L) metric (right).} The figures depict the bipartite matching between the detections (\textcolor[HTML]{66B132}{\textbf{green}}) and the ground truth objects (\textbf{black}). Regular 3D\,AP matching (left) is based on the intersection over union~(IoU) values and cannot match the detections that suffer from longitudinal localization errors, though the detection is reasonable and can provide useful signals to down stream modules. In contrast to this, the proposed LET-3D-AP(L), shown on the right, is more permissive by shifting the predictions to mitigate the longitudinal localization errors. We show the shifted predictions in \textcolor[HTML]{00A5E6}{\textbf{blue}}, which are used for computing the longitudinal error tolerant intersection over union (LET-IoU). To account for the used longitudinal tolerance, we propose longitudinal affinity (LA) as a measure of how close the original prediction is to the ground truth in the longitudinal direction.}
\label{fig:teaser}
\end{center}%
}]

\thispagestyle{empty}
\pagestyle{empty}

\begin{abstract}
The 3D~Average Precision (3D\,AP) relies on the intersection over union between predictions and ground truth objects. However, camera-only detectors have limited depth accuracy, which may cause otherwise reasonable predictions that suffer from such longitudinal localization errors to be treated as false positives. We therefore propose variants of the 3D\,AP metric to be more permissive with respect to depth estimation errors. Specifically, our novel longitudinal error tolerant metrics, LET-3D-AP and LET-3D-APL, allow longitudinal localization errors of the prediction boxes up to a given tolerance. To evaluate the proposed metrics, we also construct a new test set for the Waymo Open Dataset, tailored to camera-only 3D detection methods. 
Surprisingly, we find that state-of-the-art camera-based detectors can outperform popular LiDAR-based detectors with our new metrics past at 10\% depth error tolerance, suggesting that existing camera-based detectors already have the potential to surpass LiDAR-based detectors in downstream applications.
We believe the proposed metrics and the new benchmark dataset will facilitate advances in the field of camera-only 3D detection by providing more informative signals that can better indicate the system-level performance.
\footnote{Work done while at Waymo.}
\end{abstract}

\section{Introduction}
\label{sec:intro}

Detecting objects in 3D space is a fundamental task in many robotics applications, including autonomous driving, unmanned aerial vehicles, robot navigation, and  augmented reality. While LiDAR-based object detection~\cite{sun2021rsn,shi2020pv, lang2019pointpillars,meyer2019lasernet,qi2019deep,zhou2018voxelnet,sindagi2019mvx} has been studied extensively in recent years with impressive results, reliable camera-based 3D object detection remains a challenging and active area of research~\cite{chen2016monocular,brazil2019m3d,mousavian20173d,xu2018multi,simonelli2019disentangling,zhou2019objects,wang2021fcos3d,reading2021categorical,wang2019pseudo,wang2022detr3d,roddick2018orthographic,rukhovich2022imvoxelnet,cramnet2022hwang}.

According to common metrics, LiDAR-based detectors outperform camera-based detectors by a large margin, suggesting that camera-based object detectors fail to detect many objects. However, a careful evaluation of the failure cases reveals that many camera-based detectors identify objects reasonably well. The issue is rooted in these detectors often poorly estimating depth. Common metrics, such as the 3D Average Precision~(3D\,AP), rely on the intersection over union~(IoU) to associate the prediction boxes with the ground truth boxes. As a result, they may treat reasonable predictions that suffer from these longitudinal localization errors as false positives, leading to lower and less informative scores.

We therefore propose a variant of the 3D\,AP metric that is more permissive to depth errors. The proposed longitudinal error tolerant metrics, LET-3D-AP and LET-3D-APL, allow longitudinal localization errors up to a given tolerance. Specifically, we define the longitudinal error to be the object localization error along the line of sight between the camera and the ground truth object. The maximum longitudinal error that our metric tolerates is an adjustable percentage of the distance between the camera and the ground truth object.

For each prediction and ground truth pair with tolerable longitudinal error, we correct the longitudinal error by shifting the prediction box along the line of sight between the camera and the center of the prediction box. We then use the resulting corrected box to compute the IoU, which we refer to as the longitudinal error tolerant IoU (LET-IoU). The precision and recall values are then computed by performing a bipartite matching with weights based on the longitudinal affinity and the LET-IoU values.

Finally, we define two longitudinal error tolerant metrics, LET-3D-AP and LET-3D-APL. First, the metric LET-3D-AP is the AP on matching results using proposed method. Note that this metric does not penalize any corrected errors and is therefore comparable to the original 3D\,AP metric, but with more tolerant matching. In contrast to this, the metric LET-3D-APL penalizes longitudinal localization errors by scaling the precision.



To evaluate the proposed metrics, we extended the existing Waymo Open Dataset~\cite{sun2020scalability} by (1) creating a new dedicated test set of 80 segments (80,000 images) with LiDAR data redacted, (2) providing camera-synchronized 3D box labels for the entire dataset, eliminating the camera/label synchronization gap. Waymo included these extensions as part of the Waymo Open Dataset 3D Camera-Only Detection Challenge. The challenge remains open for submissions, and can be used to benchmark camera-only detection methods.

\section{Related Work}
\label{sec:related_work}

Following popular 2D detection benchmarks, such as PASCAL VOC~\cite{everingham2015pascal} and COCO~\cite{lin2014microsoft}, most 3D object detection benchmarks rely on the metric 3D\,AP to evaluate detections based on the intersection over union (IoU), either in 3D or in a bird's eye view (BEV), with predefined IoU thresholds. For example, the KITTI dataset~\cite{Geiger2012CVPR} and the Waymo Open Dataset~\cite{sun2020scalability} adopt 3D IoU as the main matching function. The Waymo Open Dataset proposes to use a heading accuracy weighted AP, referred to as APH, as the primary metric to penalize incorrect heading prediction. NuScenes~\cite{nuscenes2019} uses the center distance between predicted objects and ground truth objects as the true positive matching criterion and proposes a set of true positive metrics to quantify other errors, including localization, scale, and orientation.

While LiDAR-based detectors~\cite{sun2021rsn,shi2020pv, lang2019pointpillars,meyer2019lasernet,qi2019deep,zhou2018voxelnet,sindagi2019mvx} remain the most popular
methods for 3D object detection for autonomous driving, monocular camera-based 3D object detection has been gaining traction in recent years. Methods can be roughly categorized into two groups: 1) The first camp leverages mature perspective 2D detectors and equips the networks with 3D box attributes for 3D detection~\cite{chen2016monocular,brazil2019m3d,mousavian20173d,xu2018multi,simonelli2019disentangling,zhou2019objects,wang2021fcos3d,liu2020reinforced,simonelli2020towards,chen2020monopair,li2020rtm3d,wang2022mv}. An emerging direction is to leverage Transformer models for implicit depth encoding~\cite{liu2022petr,wang2022detr3d,huang2022monodtr,zhang2022monodetr,liu2022petrv2}. 2) The second camp leverages mature LiDAR detectors and projects 2D perspective features into 3D in the form of point clouds~\cite{wang2019pseudo,ma2020rethinking,you2019pseudo,ding2020learning,cramnet2022hwang,qian2020end,chen2020dsgn,guo2021liga}, feature maps in bird's eye view (BEV)~\cite{li2022bevformer,weng2019monocular,reading2021categorical,wang2019pseudo,wang2022detr3d,huang2021bevdet,li2022bevdepth,xie2022m,jiang2022polarformer,liu2022bevfusion,huang2022bevdet4d,zhang2022beverse,chen2022graph,chen2022polar,roh2022ora3d}, or in voxel space~\cite{roddick2018orthographic,rukhovich2022imvoxelnet,wang2022monocular,lu2022learning}. These methods then predict the detections by applying 3D detection heads to the 3D/BEV feature maps. Accurate monocular depth estimation is essential for all methods. Since monocular depth estimation is intrinsically an ill-posed problem, however, it is difficult for camera-based methods to accurately estimate the depth that corresponds to the objects, resulting in longitudinal localization errors that lead to reduced 3D\,AP scores when compared to LiDAR-based methods.

As pointed out by Ma \emph{et al.}~\cite{ma2021delving}, the performance of monocular camera 3D~detection methods can be greatly improved when their longitudinal localization errors are mitigated by using ground truth depth or localization. This is why we are proposing longitudinal error tolerant~(LET) metrics for evaluating camera-only 3D detection methods. While generalized intersection over union (GIoU)~\cite{Rezatofighi_2018_CVPR} and its 3D variant~\cite{xu20193d} can also be used to match non-overlapping bounding boxes, they do not target errors in a specific direction, and the shape and heading mismatch is not factored into the non-overlapping cases.

\section{Longitudinal Error Tolerant 3D AP}
The 3D\,AP metric relies on the IoU to match prediction boxes with ground truth boxes. Therefore, prediction boxes that have little or no overlap with the corresponding ground truth boxes will be treated as false positives. However, these predictions may still contribute valuable information to the decision making of an autonomous driving system. We therefore propose a metric that rewards such detections.

Our proposed metrics are inspired by the following objectives: First, given a model of an assumed localization error distribution, design a new matching criterion so that a predicted box may still match with a target ground truth box even when they do not match in terms of IoU. Second, design a new bipartite matching cost function that takes the localization, shape, and heading errors into account so that frame-level matching can be properly calculated. Last, design a penalty term to penalize detections that can only be matched with the ground truth when using the longitudinal error tolerant matching.

\subsection{Localization Errors in Camera-Based 3D Detection}
\begin{figure}[t]
    \centering
    \includegraphics[width=0.8\linewidth]{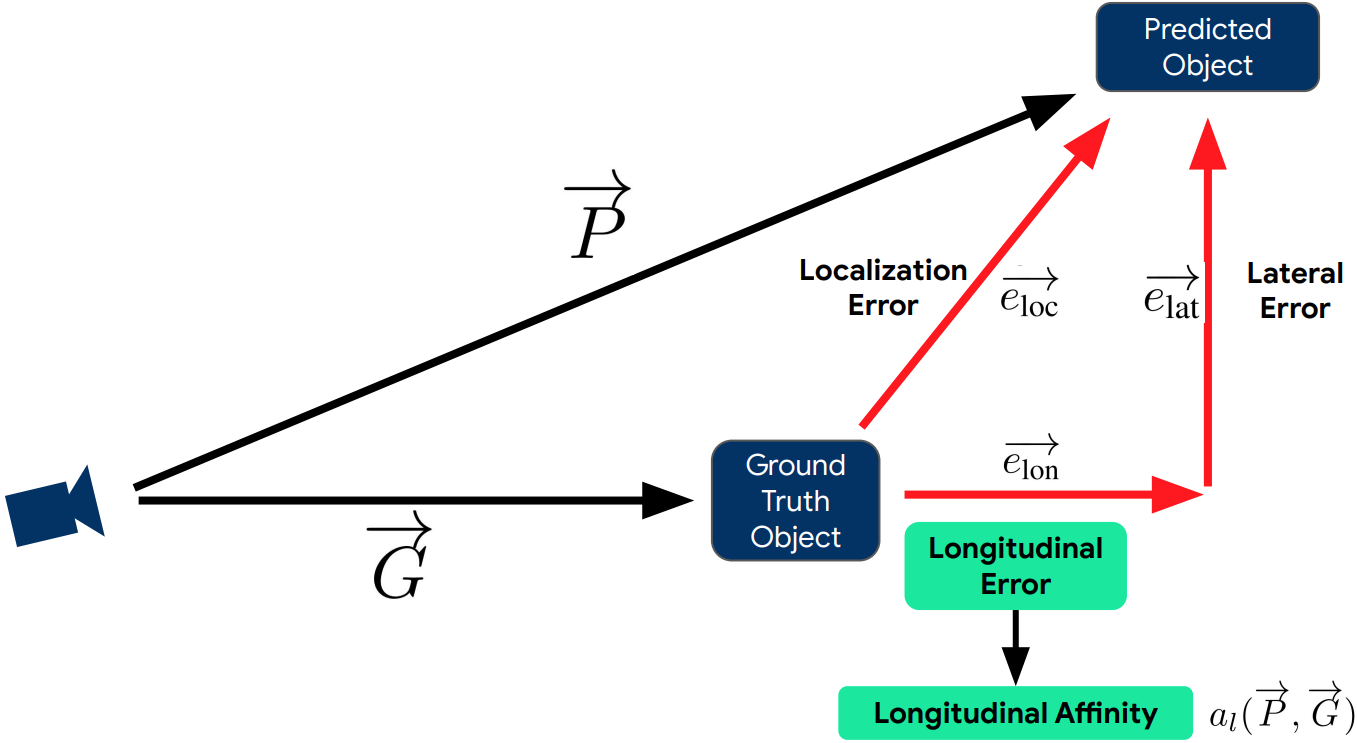}
    \caption{\textbf{Breakdown of the localization error.} We decompose the 3D detection localization error into a lateral error and a longitudinal error. We find that the longitudinal error is more prominent in camera-only 3D detection. We therefore propose longitudinal error tolerant~(LET) metrics that are more permissive with respect to the longitudinal localization error.}
    \label{fig:localization_error}
\end{figure}

We assume a ground truth bounding box $G$ with center $\vect{g} \in \mathcal{R}^3$ and a predicted bounding box $P$ with center $\vect{p} \in \mathcal{R}^3$ such that the origin $[0,0,0]$ is the camera location, or a mean position in the case of multi-camera setup. That is, all 3D location vectors are in the camera frame. We define the localization error as the vector
\begin{equation}
    \vect{e}_{\text{loc}}=\vect{p}-\vect{g}.
\end{equation}
We decompose the localization error into two components:
\begin{itemize}[noitemsep]
    \item The longitudinal error $\vect{e}_{\text{lon}}$ is the error along the line of sight from the center of the prediction box to the center of the ground truth box, giving $\vect{e}_{\text{lon}}=(\vect{e}_{\text{loc}} \cdot \vect{u}_G) \cdot \vect{u}_G$, where $\vect{u}_G = \vect{g}/|\vect{g}|$ is the unit vector of $\vect{g}$.
    \item The lateral error $\vect{e}_{\text{lat}}$ is the distance between the predicted box~$P$ and the line of sight to the ground truth box~$G$. It is defined as the shortest distance from the predicted box to any point on the line of sight, leading to $\vect{e}_{\text{lat}}=\vect{e}_{\text{loc}}-\vect{e}_{\text{lon}}$.
\end{itemize}

Figure~\ref{fig:localization_error} illustrates the error terms. We observe that localization errors tend to have the following attributes for camera-based 3D detectors. Localization errors tend to be the most pronounced along the line of sight because of imperfect depth estimation. We assume that the standard deviation of the longitudinal error, $\vect{e}_{\text{lon}}$, is proportional to the distance between the sensor and the center of the ground truth bounding box, $\vect{g}$. The lateral error, $\vect{e}_{\text{lat}}$, is the result of an imperfect estimation of the object center on the camera image plane. Since the size of the bounding box projected onto the camera plane is inversely proportional to the range to a given ground truth, $G$, the standard deviation of the center estimation error in pixels, $\sigma(|\vect{e}_{\text{cam}}|$), is also inversely proportional to the range of a given ground truth $\vect{g}$, that is,  $\sigma(|\vect{e}_{\text{cam}}|) \propto 1/|\vect{g}|$. The lateral error in 3D space, however, is scaled by the range projection, that is, $\vect{e}_{\text{lat}} =|\vect{g}| \cdot \sigma(|\vect{e}_{\text{cam}}|)$. Therefore, the standard deviation of the lateral error is independent of the range of the ground truth object and can be set as a constant.
We therefore only introduce a tolerance for the longitudinal localization errors.
                                                          
\subsection{Longitudinal Affinity}
\begin{figure}[t]
    \centering
    \includegraphics[width=0.8\linewidth]{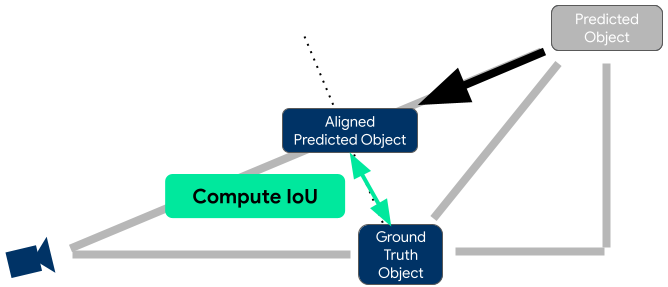}
    \caption{\textbf{Computing LET-IoU.} Given a predicted object and a ground truth object to be matched with, we move the predicted object along the line of sight to obtain minimal distance to the ground truth center. We then compute the LET-IoU as the 3D-IoU between the aligned predicted object and the ground truth object.}
    \label{fig:p_aligned}
\end{figure}

We propose a scalar value, longitudinal affinity $a_l(\vect{p}, \vect{g})$, to determine the scores for matching prediction boxes with ground truth boxes given a tolerance for the longitudinal error. Specifically, the longitudinal affinity, whose value is in $[0.0, 1.0]$, estimates how well the centers of a prediction box and the ground truth box align. Given the longitudinal error, $\vect{e}_{\text{lon}}$, between a pair of prediction and ground truth, we define the longitudinal affinity based on the following hyperparameters:
\begin{itemize}[noitemsep]
    \item \variable{longitudinal\_tolerance\_percentage} $T_l^p$: The maximum longitudinal error $\vect{e}_{\text{lon}}$ is expressed as a percentage $T_l^p$ of the range to the ground truth~$G$. For example, $T_l^p=0.1$ provides a 10\% tolerance, and for a ground truth object that is 50~meters away ($|\vect{g}|=50$), the longitudinal tolerance is 5 meters.
    \item \variable{min\_longitudinal\_tolerance\_meter} $T_l^m$: When a ground truth object is close to the sensor origin, the percentage-based tolerance can result in an small matching region. This parameter controls the minimum absolute tolerance, thus mainly affecting near range objects.
\end{itemize}

Finally, we define the longitudinal affinity $a_l(\vect{p}, \vect{g})$ between a prediction box with center $\vect{p}$ and a ground truth box with center $\vect{g}$ as:
\begin{equation}
    a_l(\vect{p}, \vect{g}) =1-\min{\left(\frac{|\vect{e}_{\text{lon}}(\vect{p}, \vect{g})|}{T_{\text{l}}}, 1.0\right)},
\end{equation}
where $T_{l}=\text{max}(T_l^p  \cdot |\vect{g}|, T_l^m)$.

\subsection{LET-IoU: Logitudinal Error Tolerant IoU}
\label{subsec:let_iou}

\begin{figure}[t!]
    \centering
    \includegraphics[width=0.5\linewidth,frame]{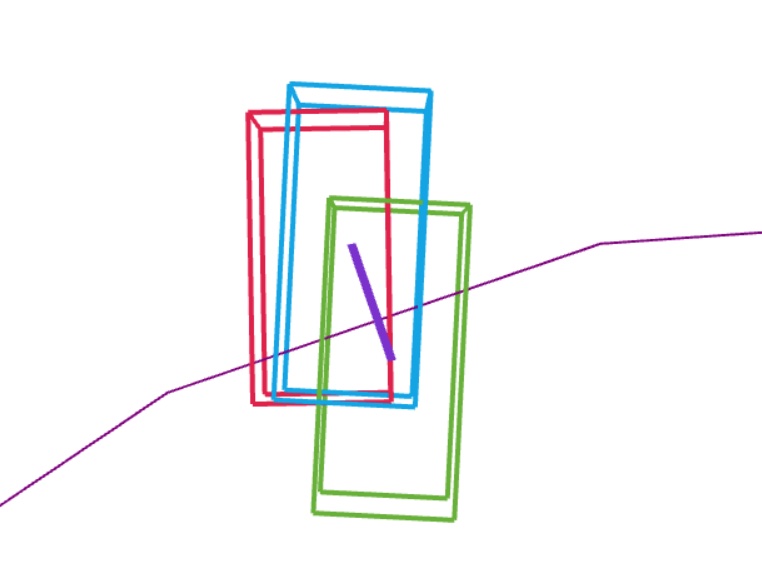}
    \caption{\textbf{An example of a matched detection using LET-IoU.} The \textcolor[HTML]{66B132}{\textbf{green}} box denotes the detection, the \textcolor[HTML]{E11845}{\textbf{red}} box denotes the ground truth, and the \textcolor[HTML]{00A5E6}{\textbf{blue}} box denotes the longitudinal aligned detection as stated in Section~\ref{subsec:let_iou}.We also show the connections between matched prediction boxes and aligned boxes using a \textcolor[HTML]{7B34CF}{\textbf{purple}} connector.}
    \label{fig:let_iou}
\end{figure}

The longitudinal affinity captures the longitudinal error of a prediction based on the center of the predicted bounding box and the center of the ground truth bounding box. To determine whether a prediction can be associated with a ground truth, we also want to take into account the shape, size, and heading error. When relying on the regular 3D\,AP, this is only captured by 3D IoU. We propose LET-IoU, where IoU is calculated between the ground truth bounding box and the prediction bounding box after compensating for the longitudinal error. Specifically, we mitigate the longitudinal error of the prediction by aligning its center along the line of sight with the ground truth.

Given a ground truth with center~$\vect{g}$ and a prediction with center~$\vect{p}$, the objective is to move the center location of the prediction box along the line of sight so that the IoU between the moved prediction box and the ground truth box is maximized:

\begin{equation}
\label{eqn:let_iou_objective}
    \vect{p}_{\text{aligned}} = \operatorname*{argmax}_{\vect{p}_{\text{aligned}}} ~\text{3D-IoU}(P_{\text{aligned}}, G),
\end{equation}
where $P_{\text{aligned}}$ is the prediction box $P$ with updated center~$\vect{p}_{\text{aligned}}$.
However, there is no closed-form solution to (\ref{eqn:let_iou_objective}), and exhaustive search along the line of sight is computationally expensive. We therefore approximate the objective by minimizing the distance between the moved prediction box center and the ground truth center, where the longitudinal error is compensated. In other words, the center of the aligned prediction is the projection of the ground truth center~$\vect{g}$ onto the line of sight from the sensor to the prediction object, leading to

\begin{equation}
\label{eqn:let_iou_approximation}
    \vect{p}_{\text{aligned}}=(\vect{g}\cdot\vect{u}_P) \cdot \vect{u}_P,
\end{equation}
where $\vect{u}_P=\vect{p}/|\vect{p}|$ is the unit vector along the line of sight to prediction center $\vect{p}$. A detailed analysis on the quality of approximation (\ref{eqn:let_iou_approximation}) can be found in the supplementary material. Finally, we obtain the LET-IoU by calculating the typical 3D~IoU between the aligned prediction box~$P_\text{aligned}$ and the target ground truth box~$G$, giving
\begin{equation}
    \text{LET-IoU}(P, G) = \text{3D-IoU}(P_{\text{aligned}}, G).
\end{equation}
See Figure~\ref{fig:p_aligned} for an illustration.
Note that for simplicity, this function does not assume that the longitudinal error is within a given tolerance. In the final metrics computation, LET-IoU will only be calculated on pairs with non-negative longitudinal affinity, where $a_l>0$.

Figure~\ref{fig:let_iou} shows that the prediction (green) is aligned to the target ground truth (red), where the 3D~IoU is computed between the aligned prediction (yellow) and the ground truth, resulting in 0.7 LET-IoU.

\subsection{Bipartite Matching with Longitudinal Error Tolerance}

To calculate the precision and recall values, detection metrics need to perform a bipartite matching between the prediction set and the ground truth set. Most bipartite matching algorithms involve computing an association weight matrix $W \in \mathcal{R}^{N_{P} \times N_{G}}$, where $N_{P}$ and $N_{G}$ are the numbers of detections and ground truth objects. The objective is to maximize the summed weights among all the matched pairs. We may take into consideration the longitudinal error, the shape error, and potentially the heading error.

The typical matching weight function in 3D\,AP computes the IoU between a prediction box and a ground truth box with an IoU threshold $T_\text{iou}$:
\small
\begin{equation}
    W(i, j) =
    \begin{cases}
    \text{IoU}(P(i), G(j)), & \text{if } \text{IoU}(P(i), G(j)) > T_\text{iou} \\
    0, & \text{otherwise}
    \end{cases},
\end{equation}
\normalsize
where $P$ and $G$ are the prediction box set and ground truth box set for a single frame, respectively.
However, in our setting, there are two affinity terms to be considered: longitudinal affinity $a_l$ and LET-IoU. As a result, we propose to calculate the bipartite matching weight where both terms are taken into account:
\small
\begin{equation}
    W(i, j) =
    \begin{cases}
   a_l \cdot \text{LET-IoU}, & \text{if }  a_l > 0  \\
   & \text{and }  \text{LET-IoU} > T_\text{iou} \\
    0, & \text{otherwise}
    \end{cases},
\end{equation}
\normalsize
where we omit the function input terms $(P(i), G(j))$ of $a_l$ and LET-IoU for simplicity.
This allows us to take the shape and heading error into account while prioritizing the detections that have higher longitudinal affinity.

After computing the matching weight matrix, we run a  bipartite matching method (Greedy or Hungarian) to compute the matching results, including matched pairs (true positives / \TP), non-matched predictions (false positives / \FP), and non-matched ground truths (false negatives / \FN).

\subsection{LET-3D-AP and LET-3D-APL}
\paragraph{LET-3D-AP: Average Precision with Longitudinal Error Tolerance.}
Once the matching results are finalized, each matched prediction will be counted as a true positive (\TP). A prediction without matching ground truth is counted as a false positive (\FP). If a ground truth is not matched with any prediction, the ground truth is counted as a false negative (\FN). Then the precision $p = |\TP|/(|\TP| + |\FP|)$ and recall $r = |\TP|/(|\TP| + |\FN|)$ can be calculated. After computing the precision and recall values for detection subsets with different score cutoffs, a PR curve can be obtained. Finally, the LET-3D-AP is calculated from the average precision of the PR curve: $\text{LET-3D-AP}=\int_0^1 p(r)dr,$
where $p(r)$ is the precision value at recall $r$.

Here, we do not penalize depth errors when computing the PR curve, even though we may have had to adjust some predictions to compensate for their depth errors. This provides a number comparable to 3D\,AP but with a more tolerant matching criterion.

\paragraph{LET-3D-APL: Longitudinal Affinity Weighted LET-3D-AP.}

Here, we penalize those predictions that do not overlap with any ground truth, that is, those which only match a ground truth bounding box owing to the above-mentioned center alignment. We penalize these predictions by using the longitudinal affinity $a_{l}(\vect{p},\vect{g})$ proposed above. To this end, we propose a weighted variant of precision values.

Though the number of true positives $|\TP|$ is used both in the precision and recall computation, they represent different ideas. In precision calculation, $|\TP_P|$ means the number of matched predictions, while in the recall calculation, $|\TP_G|$ means the number of matched ground truths. As a result, we rewrite the precision and recall as
\begin{align}
\textit{Precision}&=\frac{|{\TP}_P|}{|{\TP}_P|+|\FP|} \\ \textit{Recall}&=\frac{|{\TP}_G|}{|{\TP}_G|+|\FN|}
\end{align}
to emphasize the difference between the prediction accumulator and the ground truth accumulator.

In the precision computation, we traverse through the predictions to accumulate the \FP's and \TP's. For any unmatched prediction, it only contributes to the \FP-accumulators with $1.0$. However, for a matched prediction P 
with respect to a ground truth $\vect{g}$, it contributes to \TP-accumulator with the quantity of $a_l(\vect{p},\vect{g})$ and also contributes to the \FP-accumulator with the quantity of $1-a_l(\vect{p},\vect{g})$. Essentially, we distribute part of the \TP-accumulator to the \FP-accumulator based on the longitudinal affinity. Then, the soft \TP- and \FP-accumulators are:
\begin{align}
|{\TP}_P|&=\sum_{\textit{matched P}}a_l(\vect{p},\vect{g}) \\
|\FP|&=\sum_{\textit{matched P}}(1-a_l(\vect{p},\vect{g}))+\sum_{\textit{unmatched P}}1.0 
\end{align}

Based on the definition of soft \TP~and \FP, the soft precision can then be computed as:


\begin{equation}
    \textit{Prec}_L = \frac{|{\TP}_P|}{|{\TP}_P|+|\FP|} =\overline{a_l} \cdot \textit{Precision}
\end{equation}

\normalsize
which results in a mean longitudinal affinity weighted precision point.

For the recall computation, we propose to not weight the $\TP_G$ since  the ground truth does get matched, and there is no reason to penalize the metric twice.
As a result, the precision values are discounted by the multiplier $\overline{a_l}$, which is the average longitudinal affinity of all the matched predictions that was treated as \TP.

Finally, the LET-3D-APL can be calculated with the longitudinal affinity weighted PR curve:
\begin{equation}
\text{LET-3D-APL}=\int_0^1 \textit{Prec}_L(r)dr=\int_0^1 \overline{a_l} \cdot \textit{Prec}(r)dr,
\end{equation}
where $\textit{Prec}_{L}(r)$ is the longitudinal affinity weighted precision, and $\textit{Prec}(r)$ is the precision at recall $r$. Note that the weight $a_l(\vect{p},\vect{g})$ depends on the specific score cutoff at the PR point and therefore cannot be taken out of the integral.

\section{Camera-Primary 3D Detection Dataset}

\begin{table*}[ht]
    \centering
    \scalebox{0.85}{
    \begin{tabular}{p{3.0cm}|c|c|c|c|c|c}
        \toprule
         Dataset & KITTI~\cite{Geiger2012CVPR} & Cityscapes 3D~\cite{gahlert2020cityscapes} & Argoverse 2~\cite{wilson2021argoverse} & Lyft~\cite{lyftdata} & nuScenes~\cite{nuscenes2019} & Extended WOD~\cite{sun2020scalability} \\
         \midrule
         
         
         \# Frames & 15K & 5K & 150K & 55K & 40K & 216K \\
          
         
         \midrule
          
         \# Cameras & 2 & 2 & 9 & 7 & 6 & 5 \\
         \# Surround View & 1 & 1 & 7 & 6 & 6 & 5 \\
         
         Camera Shutter Type & Global & Rolling & Global & Global & Global & Rolling \\
         Resolution [MP] & 0.7 & 2.1 & 3.1 & 1.3 / 2.1 & 1.4 & 2.5 \\
         
         \midrule
         
         \makecell[tl]{Sync Gap [ms] \\ Cam $\leftrightarrow$ Label Center}
         &  $[-12.5,12.5]$ & $0$* & $[-1.39, 1.39]$ & $[-9.7,9.7]$ & $[-4.9,4.9]$ & 0* \\
         \midrule
         
         Label Types** & \makecell[tl]{3D Bounding Box \\ 3D Optical Flow \\ 2D Camera Box \\ 2D Panoptic Seg} & \makecell[tl]{3D Bounding Box \\ 2D Panoptic Seg} & \makecell[tl]{3D Bounding Box} & \makecell[tl]{3D Bounding Box} & \makecell[tl]{3D Bounding Box \\ 3D Semantic Seg} & \makecell[tl]{ 3D Bounding Box \\ 3D Semantic Seg \\ 3D Human Keypoints \\ 2D Camera Box \\ 2D Human Keypoints \\ 2D Video Panoptic Seg} \\
         
         
      
         \bottomrule
    \end{tabular}
    }
    \caption{\textbf{Comparison of Popular Real-World AV Datasets.} We only count fully 3D box labeled frames and define a frame as representing a collection of all respective camera images in one instance. The original Waymo Open Dataset sync gap was $[-6,7]$ms and is now effectively reduced to $0$ms. The sync gaps of other datasets depend on the horizontal object position in the image plane and falls within the given range. *Cityscapes 3D is labeled on the stereo camera imagery directly, while our \textsc{camera\_synced\_box} field is the result of our optimization method, which undoes the label center sync gap considering both egomotion and independent object motion. **Some label types are only available for a subset of the data.} 
    \label{tab:compare_datasets}
    
\end{table*}

To evaluate the proposed LET metrics, we extend the existing Waymo Open Dataset~\cite{sun2020scalability}. Our contributions are two-fold: (1) We provide a dedicated test set of 80 segments with LiDAR sensor data redacted, (2) We significantly improve camera/label synchronization throughout the dataset.  
Waymo conducted the Waymo Open Dataset 3D Camera-Only Detection Challenge\footnote{\scriptsize https://waymo.com/open/challenges/2022/3d-camera-only-detection} between April and May 2022, with LET-3D-APL being the primary metric for determining the winning submissions.




Our dedicated test set contains 80 segments in the same format as the original dataset release (20s, 10Hz, 5 cameras), for a total of 1,000 camera images per segment and 80,000 images overall. LiDAR sensor data was used during labeling to ensure high-quality 3D boxes, but redacted before the release to ensure compliance with the dataset challenge rule with regards to basing predictions solely on cameras.



The original Waymo Open Dataset has a synchronization gap of $[-6,7]$ms between camera and LiDAR sensor data~\cite{sun2020scalability}. Since the 3D object labels are based on LiDAR sensor data, this gap translates into a corresponding synchronization gap between camera and LiDAR-based labels. Motion of the ego-vehicle and independent object motion can lead to misaligned labels with respect to the camera point of view.

To address this issue, we add a new field \textsc{camera\_synced\_box}, a variant of \textsc{box}, which is adjusted to eliminate the synchronization gap to the camera that perceives it. The adjustment is based on solving an optimization problem that takes into account the camera rolling shutter, the motion of the autonomous vehicle and the motion of the object of interest. 

\begin{figure}[t!]
    \centering
    \includegraphics[width=0.49\linewidth]{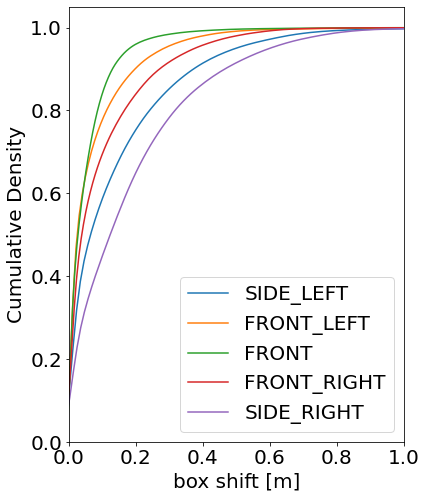}
    \includegraphics[width=0.49\linewidth]{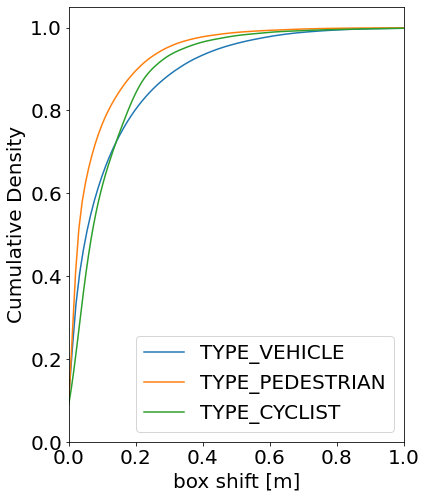}
    \caption{\textbf{CDFs of Box Shifts.} We visualize the shift between \textsc{box} and \textsc{camera\_synced\_box} based on camera types (left) and select object types (right) on the validation set.}
    \label{fig:cdf_camera_synced_box}
\end{figure}

Figure~\ref{fig:cdf_camera_synced_box} visualizes statistics of the box shift. The analysis suggests that, in practice, the shift highly depends on the camera type, with the side-facing cameras benefiting the most from improved synchronization as they are oriented orthogonally to the direction of motion and as they tend to observe close-by objects. In terms of object categories, the shift is the least pronounced for pedestrians, and the most pronounced for vehicles. This is due to pedestrians moving more slowly, reducing displacement through independent object motion, and being predominantly present in low-speed areas, reducing displacement through ego-motion.

\subsection{Comparison to Other Datasets}

Table~\ref{tab:compare_datasets} shows a comparison of popular real-world AV datasets with an emphasis on camera data and synchronization properties. Two traditional datasets in particular, KITTI~\cite{Geiger2012CVPR} and Cityscapes 3D~\cite{gahlert2020cityscapes}, are restricted by the dataset size, as well as a limited number of cameras, only spanning a comparatively small field of view. Datasets such as Lyft~\cite{lyftdata} and nuScenes~\cite{nuscenes2019} offer multiple camera views but can experience a synchronization gap of roughly $5-10$ms, depending on the object location on the image plane. For nuScenes, it is also worth pointing out that the displacement between the laser and camera sensors is more significant than in other datasets, ranging from $0.56-0.93$m. This viewpoint discrepancy makes accurate camera visibility filtering more difficult.

Both Argoverse 2 Sensor~\cite{wilson2021argoverse} and the newly extended Waymo Open Dataset are large-scale datasets that offer good synchronization. A strength of the Argoverse 2 Sensor dataset is the availability of even more surround view cameras, capturing a whole 360 degrees. A strength of the Waymo Open Dataset lies in the synergies found in many additional label modalities, including camera-based bounding boxes, human keypoints and video panoptic segmentation labels, as well as additional 3D label modalities.

\section{Experimental Results}

In this section, we analyze the proposed LET metrics against both LiDAR-based and camera-only 3D detectors. 

\begin{figure}[t!]
    \centering
    \includegraphics[width=\linewidth]{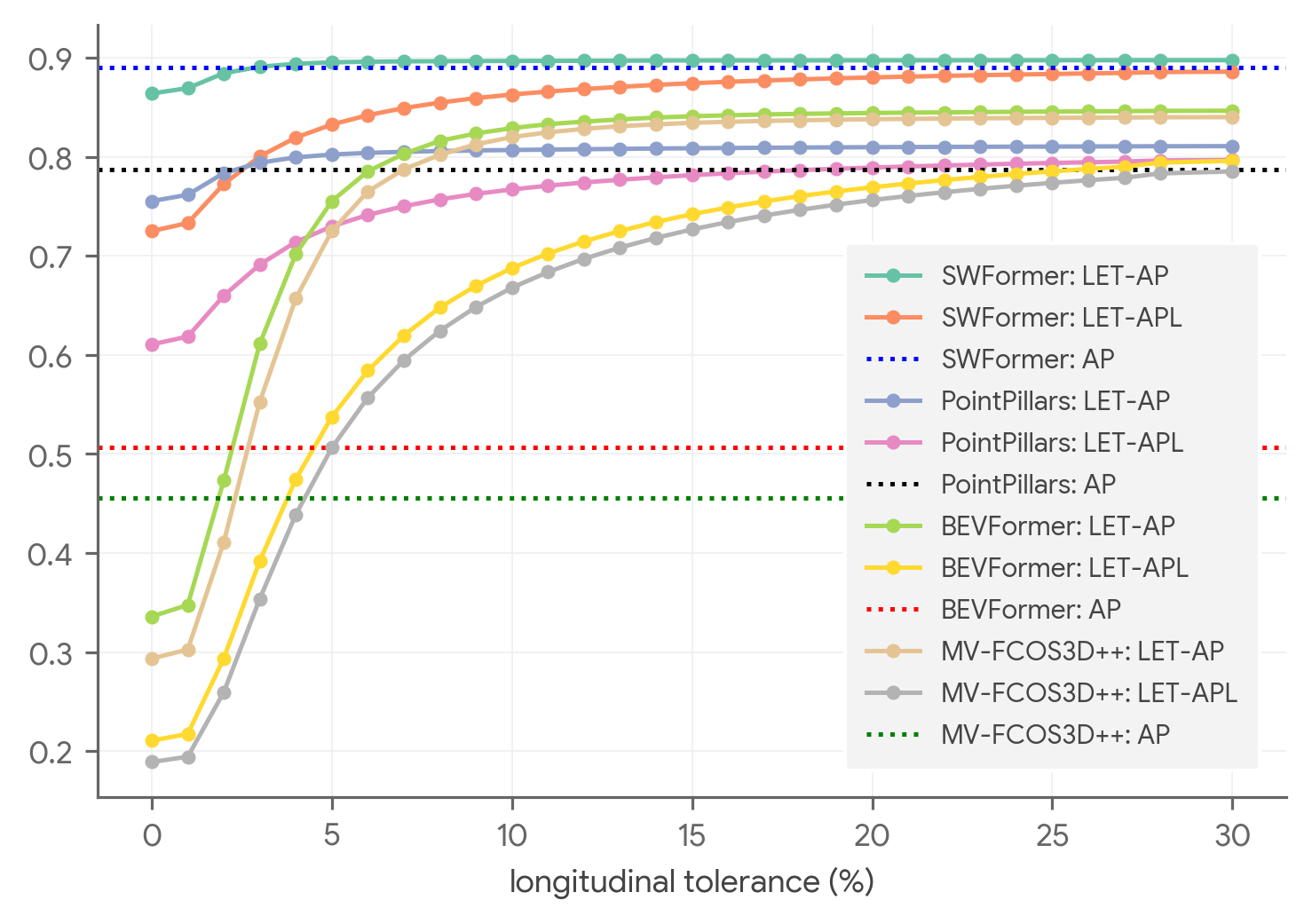}
    \caption{\textbf{LET metrics with different longitudinal error tolerance values.} We show the LET-3D-AP and LET-3D-APL, as well as the original 3D AP of the selected LiDAR-based detector (SWFormer, PointPillars) and camera-based detectors (BEVFormer, MV-FCOS3D++) on the proposed new test set. Higher tolerances lead to higher scores since more detections are being matched with ground truth objects. Surprisingly, despite BEVFormer having absolute 30\% lower 3D AP than PointPillars, it outperforms PointPillars with LET-3D-AP(L) metrics past a 10\% tolerance. This suggests that existing camera-based detectors have the potential to surpass LiDAR-based detectors, especially in downstream applications, provided that (minor) localization errors are 
    tolerated. Also, LET-3D-APL provides a smoother transition for different tolerance values and is thus more suitable for comparing camera-based 3D detectors.}
    \label{fig:tolerance_curve}
\end{figure}

\subsection{Evaluation Dataset and Baseline Methods}

To verify the proposed LET metrics, we evaluate two LiDAR-based detectors and two camera-only 3D detectors trained on the Waymo Open Dataset~\cite{sun2020scalability} (WOD) and report metrics on the newly proposed test set.

\textbf{SWFormer~\cite{sun2022swformer}}: A state-of-the-art LiDAR-based 3D detector that builds upon the idea of Swin Transformer~\cite{liu2021swin} and operates on sparse voxels. It achieves $73.36$ L2 vehicle mAPH on the WOD original test set.

\textbf{PointPillars~\cite{lang2019pointpillars}}: A well-established LiDAR-based 3D detector baseline that achieves $60.05$ L2 vehicle mAPH on the WOD original test set.

\textbf{BEVFormer~\cite{li2022bevformer}}: A camera-only 3D detector that learns unified BEV representations with spatio-temporal transformers to tackle both 3D detection and segmentation tasks. The method achieves state-of-the-art performance on nuScenes~\cite{nuscenes2019} and WOD. We evaluate a specific variant of BEVFormer that won the 2022 Waymo Open Dataset 3D Camera-Only Detection Challenge.

\textbf{MV-FCOS3D++~\cite{wang2022mv}}: A camera-only 3D detector that adds 2D supervision and temporal information onto the BEV style detector and achieves 2nd place in the 2022 Waymo Open Dataset 3D Camera-Only Detection Challenge.


\begin{table*}[t!]
    \centering
    \footnotesize
    \begin{tabular}{p{2.0cm}p{2.4cm}cccc@{\hspace{25pt}}ccc}
        \toprule
         \multirow{2}{*}{Method} & \multirow{2}{*}{Metrics} & \multicolumn{4}{c}{Class} & \multicolumn{3}{c}{Range (Vehicle)} \\
         \cmidrule(lr{20pt}){3-6} \cmidrule(r){7-9}
         \vspace{1mm}
         & & All & Vehicle & Pedestrian & Cyclist & [0, 30) & [30, 50) & [50, $\infty$)\\
         \midrule
         \multirow{4}{*}{SWFormer} & 3D\,AP (\%)      & N/A & 89.0 & 86.7 & N/A & 96.0 & 89.5 & 78.7\\
                                    & LET-3D-AP (\%)  & N/A & 89.7 & 87.3 & N/A & 96.2 & 90.2 & 79.7\\
                                    & LET-3D-APL (\%) & N/A  & 86.3 & 85.4 & N/A & 91.6 & 87.9 & 78.1\\
                                    & mLA & N/A & 0.962 & 0.978 & N/A & 0.952 & 0.975 & 0.980 \\
         \midrule
         \multirow{4}{*}{PointPillars} & 3D\,AP (\%)  & N/A & 78.6 & 58.3 & N/A & 88.2 & 76.4 & 58.4\\
                                    & LET-3D-AP (\%)  & N/A & 80.7 & 62.7 & N/A & 88.5 & 78.5 & 66.2 \\
                                    & LET-3D-APL (\%) & N/A & 76.7 & 61.0 & N/A & 83.0 & 75.9 & 54.1 \\
                                    & mLA & N/A & 0.950  & 0.973 & N/A & 0.938 & 0.967 & 0.817 \\
         \midrule
         \multirow{4}{*}{BEVFormer} & 3D\,AP (\%)     & 35.3 & 50.7 & 22.3 & 33.0 & 79.7 & 47.2 & 20.5\\
                                    & LET-3D-AP (\%)  & 70.7 & 82.9 & 71.1 & 58.1 & 89.7 & 80.6 & 50.1\\
                                    & LET-3D-APL (\%) & 56.2 & 68.8 & 53.2 & 46.5 & 75.3 & 68.5 & 38.9\\
                                    & mLA & 0.795 & 0.830 & 0.748 & 0.800 & 0.839 & 0.850 & 0.776 \\
         \midrule
         \multirow{4}{*}{MV-FCOS3D++} & 3D\,AP (\%) & 30.3 & 45.6 & 16.7 & 28.5 & 79.2 & 37.5 & 13.0\\
                                    & LET-3D-AP (\%)  & 66.0 & 82.1 & 63.0 & 53.0 & 89.3 & 78.0 & 66.1 \\
                                    & LET-3D-APL (\%) & 51.1 & 66.9 & 45.0 & 41.3 & 74.5 & 63.6 & 52.0 \\
                                    & mLA & 0.774 & 0.815 & 0.714 & 0.779 & 0.834 & 0.815 & 0.787 \\
         \bottomrule
    \end{tabular}
    \caption{\textbf{LET metrics of different breakdowns.} The methods are evaluated on the new test set with 80 multi-camera sequences. We report 3D\,AP, LET-3D-AP, LET-3D-APL, and mean longitudinal affinity (mLA) for all the class and range breakdowns, with 10\% of longitudinal tolerance. For the LiDAR-based detector, the proposed metrics show similar performance since most longitudinal tolerance is not used for matching predictions and ground truths. For camera-based detectors, the results show that proposed LET metrics are more suitable for evaluation since predictions can be better matched with ground truths, especially for small objects like pedestrian and long-range detections as indicated by the difference between 3D\,AP and LET-3D-AP(L).}
    \label{tab:challenge}
\end{table*}

\subsection{Comparisons of Longitudinal Tolerance Values}

The main hyperparameter of the proposed metrics is \variable{longitudinal\_tolerance\_percentage} $T_l^p$. We show the performance of the LiDAR-based detectors (SWFormer, PointPillars) and camera-only detectors (BEVFormer, MV-FCOS3D++) with different tolerance values in Figure~\ref{fig:tolerance_curve} for the vehicle class.

As shown in Figure~\ref{fig:tolerance_curve}, higher longitudinal error tolerance leads to higher LET-3D-AP and LET-3D-APL because more ground truth objects are matched with detections. Note that for the LiDAR-based detectors, LET-AP remains similar to the AP values for all longitudinal tolerance values since the longitudinal errors of LiDAR-based detections are already small. The results also suggest that the proposed LET metrics do not introduce additional matching errors and can reliably evaluate LiDAR-based detectors as well. 

To compute LET metrics, a longitudinal tolerance value must be chosen, and the choice of the longitudinal tolerance depends on the requirements of the downstream modules, e.g. tracking or behavior prediction, especially for an autonomous driving system. Users can also sweep the tolerance values to gain more understanding of the error patterns. 


\subsection{LET-AP v.s. AP}
We compare the proposed LET-AP and LET-APL metrics with AP on the selected baselines and show results in Figure~\ref{fig:tolerance_curve}. 
For the LiDAR-based detectors (SWFormer, PointPillars), we show that LET-AP is mostly consistent with AP, which suggests that the matching results are similar in the presence of longitudinal tolerance. For higher tolerance values, LET-3D-AP is slightly higher than 3D\,AP due to the small amount of additional detections being matched with ground truth objects with more error tolerance. Also, due to the accuracy of LiDAR range estimates, the boxes are fairly accurate, resulting in minimal difference between LET-AP and LET-APL since the allowed longitudinal tolerance is not used by most matching pairs. 

The LET-AP metrics results of the camera-only detectors in Figure~\ref{fig:tolerance_curve} exhibit a much larger gap to the AP baseline. 
This shows that many detections can be properly matched to ground truth objects with a certain amount of longitudinal tolerance, making the resulting LET-AP(L) metrics better suited for evaluating these methods. 
Under 10\% tolerance, BEVFormer achieves higher metrics than PointPillars, indicating BEVFormer may actually have better downstream performance provided sufficient robustness towards localization errors.
Compared to the LiDAR-based detectors, the drop off between LET-AP and LET-APL is larger, demonstrating higher longitudinal error among the true positive examples. 

\subsection{Metrics of Different Breakdowns}

We show detailed metrics on the proposed new test set in Table~\ref{tab:challenge}. We report 3D\,AP, LET-3D-AP, LET-3D-APL, and mean longitudinal affinity (mLA) for all class and range breakdowns, with 10\% of longitudinal tolerance. Note that the 3D\,AP metrics are computed with $0.5$, $0.3$, $0.3$ IoU thresholds for vehicles, pedestrians and cyclists.
As shown in Table~\ref{tab:challenge}, LET-3D-AP values are consistently higher than 3D\,AP values since more detections are matched with ground truth objects with additional longitudinal tolerance. The difference is especially pronounced for small objects like pedestrians, as well as far-away objects. We also report the mean longitudinal affinity values to show the difference between LET-3D-AP and LET-3D-APL. We observe that methods with higher metrics not only have more valid detections but also have better localization as indicated by higher mLA. We can also verify the assumption of depth errors being roughly proportional to range by checking the consistent mLA values with different range breakdowns. Our results suggest that, under the proposed metrics, some camera-only 3D detectors can already surpass the LiDAR-based detectors.

\section{Conclusion}
We proposed LET-3D-AP(L) metrics for evaluating camera-only 3D detectors. The tolerance along the longitudinal axis allows detections to be associated with the ground truth objects despite depth estimation errors.  The results show that state-of-the-art camera-based detectors can outperform popular LiDAR-based detectors with our metrics, suggesting that existing camera-based methods have the potential in real-world applications. We also construct a new test set for the Waymo Open Dataset, tailored to camera-only 3D detection methods. We hope the proposed metrics and dataset will help advance the field of camera-only 3D detection by providing a more meaningful indication for method performance.







{\small
\bibliographystyle{IEEEtran}
\bibliography{egbib}
}

\end{document}